\documentclass[graybox]{svmult}

\usepackage{type1cm}        
\usepackage{makeidx}         
\usepackage{graphicx}        
\usepackage{multicol}        
\usepackage[bottom]{footmisc}
\usepackage{color}
\usepackage{colortbl}
\usepackage[colorlinks]{}
\usepackage{wrapfig,lipsum,booktabs}

\usepackage{newtxtext}       %
\usepackage{newtxmath}       

\makeindex             

\begin{document}

\title*{Modeling Performance in Open-Domain Dialogue with PARADISE}
\author{Marilyn Walker, Colin Harmon, James Graupera, Davan Harrison and Steve Whittaker}
\authorrunning{Walker, Harmon, Graupera, Harrison and Whittaker} 
\institute{University of California, Santa Cruz, Santa Cruz, Ca., 95064 \email{mawalker@ucsc.edu}
}
%
%
\maketitle

\abstract{There has recently been an explosion of work
  on spoken dialogue systems, along with an increased interest in
  open-domain  systems that engage in  casual
  conversations on  popular topics such as
  movies, books and music. These systems aim to socially engage,
  entertain, and even empathize with their users. Since
  the achievement of such social goals is hard to
  measure, recent research has used dialogue length or human ratings as evaluation metrics, 
  and developed methods for  automatically calculating novel
  metrics, such as coherence, consistency, relevance and engagement. 
  Here we 
  develop a PARADISE  model for predicting the performance of Athena, a dialogue system that has
  participated in thousands of conversations with real users, while
  competing as a finalist in the Alexa Prize. We use both user ratings
  and dialogue length as metrics for dialogue quality, and 
  experiment with predicting these metrics using automatic features that are
  both system dependent and independent. Our goal is to learn a general objective function that
  can be used to optimize the
  dialogue choices of any Alexa Prize system in real time and evaluate its performance. Our best model for predicting user ratings
  gets an R$^2$ of .136 with a DistilBert model, and the best model for predicting length with system independent features
  gets  an R$^2$ of .865, suggesting that 
   conversation length may be a more reliable measure for automatic training of dialogue systems.  }

\section{Introduction}
\label{sec:intro}

Over the last ten years there has been an explosion of work on spoken
dialogue systems, along with an increased interest in open-domain
 systems that engage in casual conversations on popular topics such as movies, books and
music. These systems aim to socially engage, entertain, and even empathize with
their users \cite{paranjape2020neural,higashinaka2008effects,zhou2020design}. Since the achievement of such social goals is hard to
measure,  recent work has used dialogue length and  human ratings as evaluation metrics \cite{higashinaka2021overview,li2016deep,shalyminov2018neural}.
  Other work has focused on  automatically calculating  novel metrics such as coherence, consistency, relevance and engagement, using  supervised models,  or measures based on  
  language model probabilities and word embedding cosine similarity
 \cite{liu2016not,mehri-USR20,mehri2020unsupervised,higashinaka2019improving,pang-etal-2020-towards,bruni2017adversarial,ghazarian2020predictive,yi2019towards,yeh2021comprehensive} {\it inter alia}.
 
 

  
This paper develops a PARADISE-style dialogue evaluation  model \cite{Walkeretal97,Walker00,ultes2016analysis,ultes2017domain}, for a particular type of open-domain dialogue system, namely systems that compete in the Alexa Prize (AP) \cite{fang2018sounding18,paranjape2020neural,curry2018alana,papaioannou2017alana,harrison2020athena,patilathena2021,finch2020emora,bowden2019cui,chen2018gunrock,gabriel2020further}. 
The evaluation criteria for the AP explicitly specifies that systems will be evaluated  on a combination of ratings from real users and the length of conversations: the ``Grand Challenge'' goal is conversations that  last for twenty minutes and get an average rating of 4 out of 5 \cite{Venkatesh2017}.  Real users can initiate  AP conversations by saying {\it Let's talk} to any Alexa device. They are then randomly assigned  to an AP system, and, at  the end of the conversation, are asked for a user satisfaction rating:   {\it On a scale of 1 to 5, how interested are you in talking to this socialbot again?}. Because about 20\% of users provide  ratings, reliably predicting user satisfaction (ratings) would be valuable. Since the AP also aims for "long and engaging" conversations, conversation length is a second  measure of dialogue quality.
Conversation length makes particular sense in the context of the Alexa Prize, where real users {\bf choose} when to stop the conversation. We thus experiment
with predicting both user ratings and conversation length for Athena, a dialogue system that has been an AP finalist for the last two years \cite{harrison2020athena,patilathena2021,bowden2019slugbot}. In the 20/21 semi-finals, Athena's average overall rating was 3.62 and average length was 2.12 minutes. 

\begin{wrapfigure}{r}{2.85in}
\vspace{-.25in}
\includegraphics[width=2.85in]{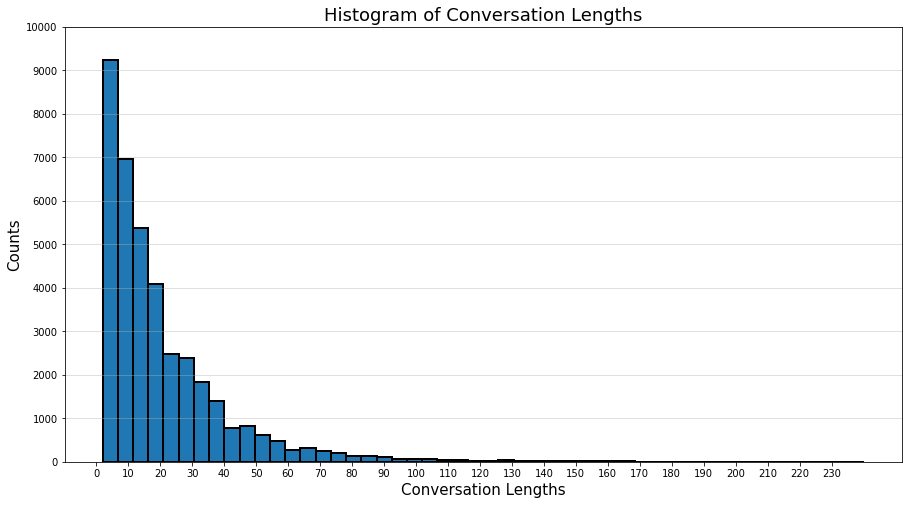}
\caption{Conversation Lengths in Exchanges} \label{conv-lengths-fig}
\vspace{-.25in}
\end{wrapfigure}
Previous  work on open-domain dialogue has been trained on
large-scale, unconstrained, freely available corpora, such as Twitter \cite{Ritteretal11,Charrasetal16,Duplessisetal16}, Reddit \cite{dodge2016evaluating}, Open Subtitles \cite{hori2017end}, and Film Scripts \cite{serban2016building,walker2012annotated}. 
More recent models are trained on controlled, crowd-sourced datasets, which are
shorter and text-based to facilitate collection, training and evaluation,  with lengths of 2-4 exchanges  for Empathetic Dialogues\cite{rashkin2019towards},  4 exchanges for Daily Dialogue\cite{li2017dailydialog},   6 exchanges for PersonaChat\cite{zhang2018personalizing},  and 10 exchanges for Topical Chat\cite{gopalakrishnan2019topical}.
Conversations with AP systems are much longer: in recent challenges conversations average more than 20 exchanges in length, with some  as long as 200 exchanges.\footnote{We use the term exchanges, also known as adjacency pairs, since in AP  parlance, a turn consists of both a user and system utterance, while a turn is a single user or system utterance in the typical usage in the spoken dialogue community.} AP conversations are also a very different genre, due to the requirement that  AP systems must
 carry out  real-time, spoken conversations with hundreds of thousands of users, and  respond to any  topic that the user might bring up, including recent events.
AP systems are expected to recognize and use up-to-date knowledge  about sports and athletes, movies and TV shows and their actors, characters and directors, as well as books and their authors, and music, bands and musicians \cite{gabriel2020further,shang2021entity,bowden2018slugnerds}. Thus,  AP systems require  mechanisms beyond  training on  static open-domain dialogue corpora  \cite{Ritteretal11,zhang2020dialogpt,brown2020language,radford2019language,hedayatnia2020policy}.

Section~\ref{corpus-sec} describes our corpus   of  Athena dialogues.  Section~\ref{methods-sec} describes our experimental setup, and Section~\ref{results-sec} presents both quantitative and qualitative results. We discuss
related work throughout the paper where it is relevant, 
and sum up our results and future work in Section~\ref{conc-sec}. We show that the best model for predicting user ratings results in an R$^2$ of 0.135, while   a model predicting conversation length using Athena Independent features results in an  R$^2$ of .862.  These results suggest that conversation length could be a more reliable indicator of dialogue quality in  large scale open-domain dialogue corpora that have been collected using interactions with real users.
We expect these results to generalize to other AP  systems and be useful for  optimizing the system's dialogue policy using reinforcement learning. 

\section{Athena Dialogue Corpus}
\label{corpus-sec}

\begin{wrapfigure}{r}{1.5in}
\vspace{-.25in}
\includegraphics[width=1.25in]{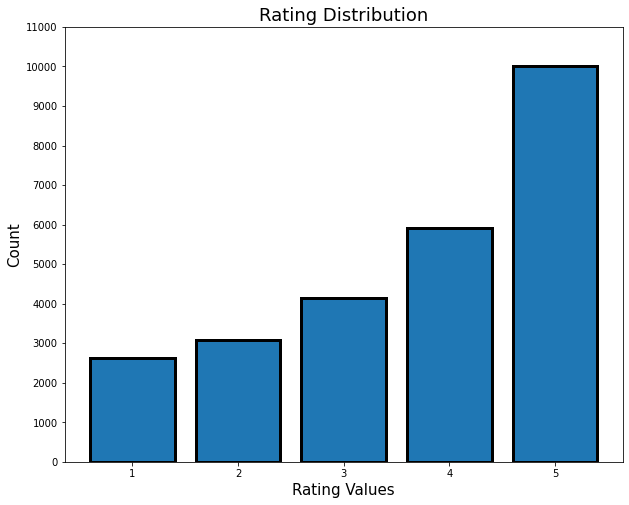} 
  \caption{Conversation Ratings} \label{ratings-fig}
  \vspace{-.2in}
\end{wrapfigure}
We sampled a corpus of $\sim$32K  rated Athena  dialogues from 2021.
Figure~\ref{conv-lengths-fig} shows that many conversations are very short, often only consisting of one exchange, where the  user seems to have invoked the AP  skill by accident. The average conversation length is around 20 exchanges, with some conversations  as long as 200 exchanges. The conversation ratings in  Figure~\ref{ratings-fig} shows that highly rated conversations dominate, with a median of 4 and an average of  3.7. 

\begin{wraptable}{r}{1.75in}
\vspace{-.25in}
  \small
    \begin{tabular}{p{1.1in}|r}
    \toprule
        Metrics & Correlation  \\ \hline \hline
        Rating/Length & 0.134**   \\ \hline 
        Rating/Compliments & 0.068** \\ \hline %
        Rating/Complaints &  -0.074** \\ \hline 
        Length/Compliments & 0.216** \\ \hline 
        Length/Complaints & 0.022** \\ \bottomrule 
    \end{tabular}
  \caption{Correlations between metrics: ** marks correlations  $p \leq .01$.}
  \label{table:correlation-stats}
  \vspace{-.2in}
\end{wraptable}
We expected a strong correlation between length and rating
but surprisingly, Row 1 of Table~\ref{table:correlation-stats} shows that these two metrics are only weakly (but significantly) correlated.
We also examine the correlations of  dialogue acts indicating user satisfaction (compliments) or dissatisfaction (complaints), illustrated in Figure~\ref{fig:compliments-complaints} and used as features in Section~\ref{methods-sec}. While all correlations are significant,  the largest is between length and compliments, similar to findings in previous work on AP dialogues\cite{shalyminov2018neural}. 
User utterances such as these have also been used as the basis for open-domain evaluation models such as FED, based on contextualized queries to DialogGPT for language model probabilities  \cite{mehri2020unsupervised,zhang2020dialogpt}.

 Figure~\ref{short-convs-fig}  provides  examples of  short conversations.\footnote{In accordance with the AP  Challenge rules, the shared  conversations 
in Figure~\ref{short-convs-fig}, Figure~\ref{fig:good-conv}
and Figure~\ref{fig:bad-conv} are between Athena and UCSC undergraduates rather than real users.} 
Each row counts as an exchange, so the first conversation is of length 3 (it includes Exchange 2 
 where the user says {\it alexa stop}). The second conversation is length 5. A conversation of length 1 only includes the greeting in Exchange 0. We believe that conversations of length$<$5 don't justify any rating at all, and that understanding how dialogue behaviors affect ratings requires  modeling the ratings  of users who experienced enough of a conversation to judge its quality.  We therefore only model conversations that are at length$\geq$ 5, since greeting the user and asking their name takes $\sim$4 exchanges. See Figure~\ref{short-convs-fig}. The final corpus has a median length of 17 exchanges and an average length of 23.7 exchanges.

\begin{wrapfigure}{r}{2.45in} 
\centering
\begin{small}
\vspace{-.2in}
    \begin{tabular}{p{1.0in}|p{1.3in}}
     \toprule
     {\bf Compliments} & {\bf Complaints} \\ \hline
     you are the best a.i. &  none of your business  \\
     you talk really nice  & you ask a lot of questions\\
     
     you're interesting  & i don't care \\
     that's so cool  & what is wrong with you\\
     i never knew that & this is pretty boring  \\  \hline
\end{tabular}
\end{small}
\vspace{-.1in}
\caption{Example Compliments and Complaints} \label{fig:compliments-complaints}
\vspace{-.2in}
\end{wrapfigure}
Figure~\ref{fig:good-conv} and Figure~\ref{fig:bad-conv}
illustrate  Athena conversations.
Sub-dialogues are logged by topic and response generator and most conversations consist of multiple topics, some initiated by Athena, and others by the user. Each topic uses different types and amounts of content, and  supports different conversation styles. Athena  supports ``niche'' topics such as astronomy, nutrition, and comic books along with popular topics such as music, animals, movies and video games\cite{patilathena2021}.
Figure~\ref{fig:good-conv} shows a sub-dialogue about {\it comic books}. Figure~\ref{fig:bad-conv} shows a sub-dialogue about  {\it movies}.

\begin{figure}[t!hb]
\begin{small}
    \begin{tabular}{@{}p{0.15in}|p{0.30in}|p{.65in} |p{3.3in} @{}}
     \toprule 
        {\bf Ex}& {\bf Topic} & {\bf User}& {\bf Athena} \\\hline \hline
        0 & intro & chat with me &	Hi, this is an Alexa Prize Socialbot. I hope you had a good weekend. How's it going. \\ \hline
        1 & intro & good & Okay, can't wait to get to know a thing or two about you. I don't think we've had a chance to meet yet. Would you like to tell me your name?	\\ \hline
        2 & & alexa stop &  \\\hline \hline
        0 & intro&	talk to me & Hi, this is an Alexa Prize Socialbot. I hope your week is off to a good start. How are you doing? \\\hline
 1	& intro & good & Okay, can't wait to get to know a thing or two about you.</amazon:emotion> I don't think we've had a chance to meet yet. Would you like to tell me your name?	\\\hline
	2	& intro & colin	&	Okay, got it. It's nice to meet you, Colin! Did you watch the fireworks on Sunday?\\\hline
	3 & intro &	yes	& Awesome! Unfortunately I couldn't go out and barbecue because I'm stuck in the cloud. But I hope you were able to celebrate! Anyway, What are some of the things you like to do in your free time. \\\hline
	4 &  intro &	stop & \\\hline
\end{tabular}
\end{small}
\caption{Two short conversations: the first (length 3) was rated  5, the second (length 5) was rated 3. Conversations of length $\geq$ 5 are included in our training data while those of length $\le$ 5 are not.}
\label{short-convs-fig}
\vspace{-.2in}
\end{figure}

\begin{figure}[htb]
    \centering
\begin{small}
    \begin{tabular}{@{}p{0.15in}|p{0.35in}|p{.99in} |p{2.9in} @{}}
     \toprule
        {\bf Ex.} & {\bf Topic} & {\bf User} & {\bf Athena}\\\hline \hline
       18 & comics& oh really &  I am familiar with DC and marvel. Who is your favorite character? \\\hline
       19 & comics & captain marvel &  What about captain marvel do you like?  \\\hline
 20	&comics &how she's like super empowering&	In general, why are you interested in comic books? \\ \hline
 21 &comics &	because they're fun to read	& That makes sense. I love comic books because they take you into another world with many different adventures. How often do you read them? \\\hline
 22&comics 	&i probably read one every week	& Okay. I wish I had more time to read comics. Would you say you're an MCU expert? Like, I mean, have you seen most of the movies? \\\hline
 23&comics 	&yeah all of them &	There's a ton of cool marvel characters in all the movies, it's hard to pick a single favorite character. How about Thor? I think Chris Hemsworth did a really good job playing him in the MCU. What do you think? Do you also think he did a good a job? \\ \hline
24&comics 	&i also think he did a good job	& Yeah! Chris Hemsworth has been playing Thor for the past 8 years in 8 different marvel movies. Don't you find that a little bit intriguing? \\ \hline
 25	&comics & i think it is &	My thoughts exactly. Here's a little trivia about the actual comic book character. Thor has actually been in 7187 different comic books. \\\hline
26&comics 	& i did not know that	& I was wondering. Tell me. If you could pick a character from DC to crossover into the Marvel Cinematic Universe, who would you pick and why? \\\hline
27&comics &	yes scarlet witch &	Yeah, that makes sense, I would pick the iconic hero Superman. The DC movies always make him look drab, but the Marvel movies are more colorful, so Superman would fit right in. Anyhow, that's enough from me. I'm interested in your opinion. If you could have any one superpower from any superhero whatsoever, what would it be and why? \\\hline
 28&comics 	& scarlet witch is power because she can live things and do things with her mind &	Yeah, I get that. I would definitely have the ability to fly like Superman. Being able to just take off at light speed and get anywhere I want in no time would be amazing. Anyway,  I think it's cool that marvel comics has been around for so long. Marvel published a Generic Comic Book, in order to trademark the names "Super-Hero" and "Super-Villain" \\ \hline
 29 &comics  &	i know that	DC is one of the biggest and most famous comic distributors out there. &  Did you ever wonder if DC actually stands for anything? It actually stands for Detective Comics, so when you say "DC Comics," you're really saying Detective Comics Comics. \\ \hline
30&comics 	&that's pretty funny &	... \\ \hline
    \end{tabular}
\end{small}
    \caption{Sub-dialogue on the {\it comics book} topic of a rated 5 conversation totaling 41 exchanges.}
    \label{fig:good-conv}
\end{figure}

\begin{figure}[h!tb]
    \centering
\begin{small}
    \begin{tabular}{@{}p{0.25in} | p{0.35in}|p{.6in} |p{3.2in} @{}} \toprule 
    {\bf Ex}& {\bf Topic} & {\bf User}& {\bf Athena} \\
    \hline \hline
    36 & movies & no i have one & I really like the Actor Eddie Murphy. Have you seen any of his movies?\\ \hline
    37 & movies & no & I really like the Movie Black Panther. Have you seen that movie?\\ \hline
    38 & movies & no & Scarlett Johansson did a really good job in her role in the Avengers. Scarlett Johansson wanted the role of Black Widow so much that she dyed her hair red to convince producers that she was right for the part.\\ \hline
    39 & movies & i don't care & Black Panther has actors David S. Lee and Winston Duke in it. What's your opinion of their performances?\\ \hline
    40 & movies & i don't care & I believe I read that Black Panther is directed by Ryan Coogler. Sometimes a director's unique voice really comes through in their work, do you think that happened in this case?\\ \hline
   \end{tabular}
\end{small}
\caption{Sub-dialogue on the {\it movies} topic of a rated 1 conversation  totaling 41 exchanges. Figure~\ref{fig:compliments-complaints} includes ``I don't care'' as a complaint.}
\label{fig:bad-conv}
\end{figure} 
    
\begin{figure}[t!hb]
\begin{center}

\includegraphics[width=4.6in]{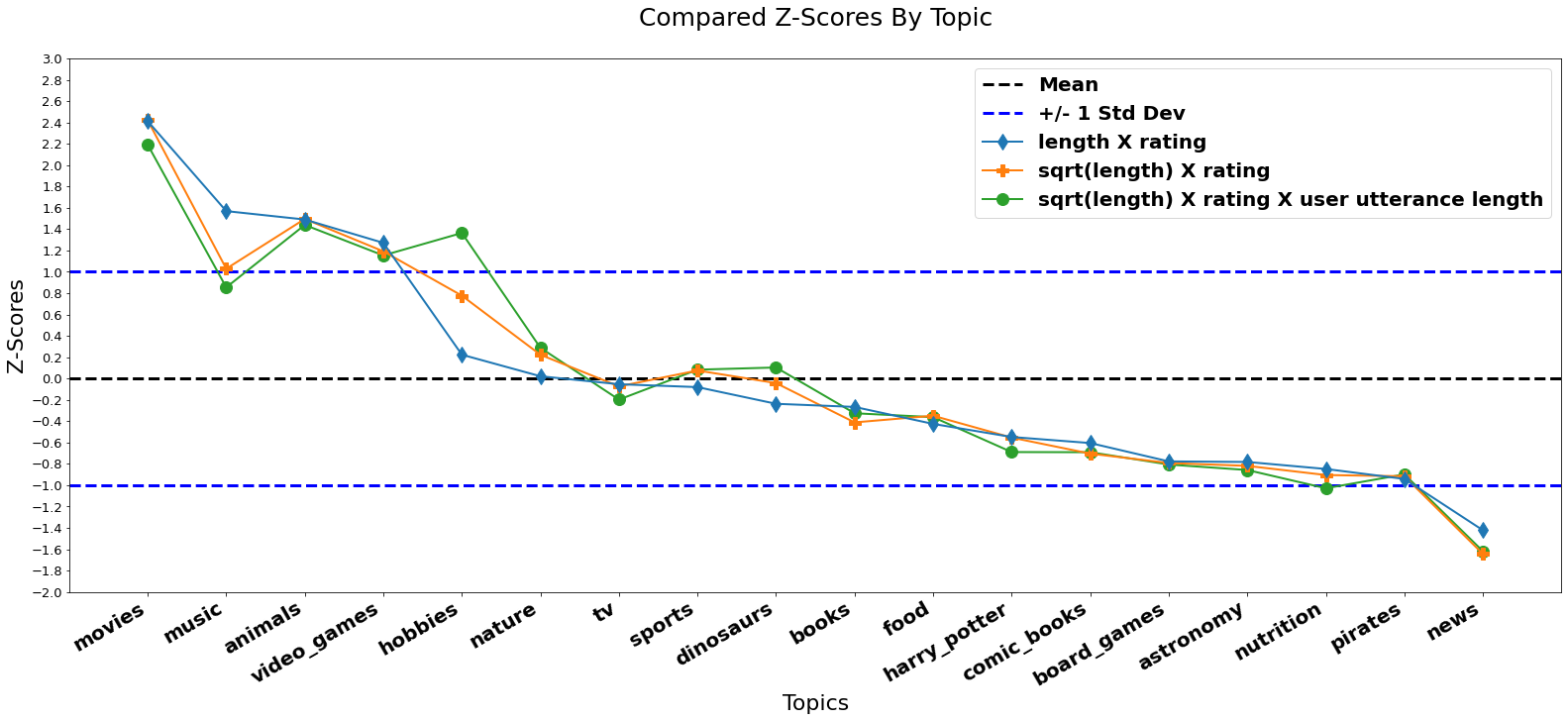} 
\vspace{-.1in}
  \caption{Z-scores of Ratings by Topic, for 3 different scoring functions} \label{z-scores-fig}
\vspace{-.2in}
\end{center}
\end{figure}
A qualitative analysis of the goodness of topical sub-dialogues suggests that topic alone should have a large effect on user ratings. However, each conversation only has one rating. We approximate topic ratings with three heuristic  scoring functions  that take into account  the  conversation rating, the length of the sub-dialogues for each topic, and the length of user utterances in topical sub-dialogues,  assuming that  longer user utterances indicate greater engagement.  For each function, we create a population of scores over all conversations and then sum them and standardize them by calculating their Z-scores (standardized value).
%
Figure~\ref{z-scores-fig} shows the Z scores for each  topic for each  scoring function. The first function (in blue in Figure~\ref{z-scores-fig}) simply multiplies the number of exchanges on a topic by the rating of that conversation, and then sums the scores over the whole population of conversations. The second function (in orange in Figure~\ref{z-scores-fig}) down-weights the effect of the length of the conversation by taking the square root of the length and then multiplying by rating. The third function multiplies each product by an additional factor of average user utterance length (in green in Figure~\ref{z-scores-fig}). The relative ranking of topics changes little over the three  functions, but hobbies, which has shorter conversations, but longer user utterances, moves from 5th place to 2nd when user utterance length is taken into account. Despite the shorter conversations,  users appear to find the hobbies topic, where Athena discusses what they like to do in their free time, very engaging.

\section{Experimental Setup}
\label{methods-sec}
Our dataset consists of  32,235 conversations split ($\sim$80/10/10) into training (25,794), development (3,229), and test   (3,212). Our goal is to develop models capable of predicting the goodness of a conversation using
regression models trained to predict either user ratings or conversation lengths.\footnote{We carried out pilot experiments on binary classifiers by treating  conversations rated 1 and 2 as bad, and conversations rated 5 as good, but these barely performed above the baseline.} 
We also experiment with predicting 
dialogue outcomes in terms of length by only using features from  the first 10 or 15 utterances in the conversation,
 to investigate whether we can make reliable predictions as the conversation unfolds, which could be then used as state variables for dialogue policy optimization, as in  previous work on {\sc problematic dialogue} prediction \cite{HastieetalACL02,LWRGL99,WLHWG02}, 
Given the AP  goal of ``long and engaging'' conversations, a problematic dialogue is one that is short.

The plot of conversation ratings in Figure~\ref{ratings-fig} has a mean of 3.7 and a median of 4.
Figure \ref{conv-lengths-fig} shows a long tail for conversation lengths, with  only 3\% of conversations 
longer than 75 exchanges(2.52 standard deviations above the mean of 23.7). We represent these longer conversations as if they had length 75.


\begin{wraptable}{r}{2.5in}
\centering
\vspace{-.2in}
\small
\begin{tabular}{p{.75in}|p{1.6in}}
    \hline
    {\bf User Utterance} & LengthMedian, Abuse\%, SDA\%  MIDAS\% Repeat\%, Red\% \\ \hline
    {\bf Topic} & Topic\%, RG\%, Topic\_Median \\ 
    \hline
\end{tabular}
\caption{Extracted Features by Category.  
}
\label{tab:features}
\vspace{-.3in}
\end{wraptable}
We develop features as real-time automatic proxies for  Athena's performance that are specific to Athena as well as dialogue quality features that we expect to generalize, as summarized in
Table~\ref{tab:features}  \cite{Walker00,Singhetal00,LitmanNAACL00}. We use  frequencies to ensure that features don't indirectly encode conversation length.
Previous work on PARADISE  included many such features, e.g. Reprompts counted system utterances where a question was repeated, and Apologies counted system utterances apologizing for misunderstanding\cite{WalkerPassonneau01,Walkeretal02b}.
 We standardize each feature with their Z-scores so their distribution  has a mean of 0 and a standard deviation of 1.
 This prevents the relative scales of features from impacting model performance  and, for linear models, indicates the relative importance of each feature with the magnitude of the weight assigned to it. 


\noindent{\bf User Utterance Features}.
To represent user utterances we calculate the median number of words in each response (LengthMedian in Table~\ref{tab:features}), expecting longer responses to indicate greater engagement. To capture user behaviors at a general level, we use two types of dialogue act taggers \cite{WalkerPassonneau01}.
First, we utilize all the MIDAS dialogue act tags (MIDAS\%) from Athena's NLU \cite{harrison2020athena,yu2021midas}: these are  used by the dialogue manager to condition Athena's conversational behaviors. These include DAs identifying the user responding negatively to a question or criticizing the system. 

Athena also uses a second level of fine-grained dialogue acts. These Social Dialogue Acts (SDAs) identify specific actions, feelings, and intents in the user's speech. Some SDAs  are grouped as negative (SDA\_complaint) or positive feedback (SDA\_compliment),   as shown in Figure~\ref{fig:compliments-complaints}. Rows 2 to 4 in Table~\ref{table:correlation-stats} show these independent measures of dialogue quality have 
a lower correlation with ratings and a higher correlation with length.  
Other SDAs, such as SDA\_dev\_command, identify
cases where the user request can only be satisfied outside the AP skill,  such as requests to play some music or to turn down the volume. These requests result in an Apology from Athena.  
Other features count dialogue acts where users ask Athena to repeat herself  (Repeat\%), and where user abuse Athena or engage in profanity (Abuse\%), which may correlate with lower ratings. We also create features measuring user utterances
on prohibited topics of conversation (Red\%), which result in Athena saying  {\it I am not the best person to discuss that with}.

\noindent{\bf Topic Features}. Some users would be expected to inherently find certain topics more interesting and the analysis in Section~\ref{corpus-sec} suggests  Athena's performance is better on some topics.   Thus, we include Athena dependent features representing the fraction of the conversation spent on each topic (e.g. TOPIC\_FREQ\_movies), as well as the median number of exchanges over all topics (TOP\_DIST\_MEDIAN). A high median indicates that the user found most topics engaging.

Figure~\ref{fig:good-conv} and Figure~\ref{fig:bad-conv} in Section~\ref{corpus-sec}
provide examples of Athena conversations that illustrate several of these features.
For example,  exchanges 18 and 26 in Figure~\ref{fig:good-conv} shows the user saying {\it Oh really} and {\it I didn't know that}, which are classified as compliments and contribute to  the SDA\_compliments\ for that conversation. Even though {\it comic books} is  one of the less highly rated topics in Figure~\ref{z-scores-fig}, this user is engaged, contributing long utterances, which are captured by the feature LengthMedian. The dialogue was 41 exchanges long, so the 13 comic exchanges here are represented by the feature Topic\_Freq\_comics having the value of  13/41.
Figure~\ref{fig:bad-conv} shows a sub-dialogue on the {\it movies} topic. Despite {\it movies} being Athena's highest rated topic (see Figure~\ref{z-scores-fig}), this user is not engaged. Exchanges 37 and 38 illustrate negative answers from the user (MIDAS\_neg\_answer), and exchanges 39 and 40  illustrate complaints. The LengthMedian  is low due to the many short user utterances. The TOPIC\_FREQ\_movies value is 5/41.

\noindent \textbf{Model Setup.}
We train regression models  using Support Vector Regression models (SVR), Decision Tree and Random Forest models, Multi-layer Perceptron models (MLP), and Transformers. Our models are implemented using the Scikit-Learn Python package~\cite{scikit-learn} and Huggingface Transformers.

\begin{itemize}
    \item \textbf{Linear Regression.} Most  linear regression experiments were done using ordinary least squares linear models. We also experimented with ridge and lasso regression models but they either matched or under-performed the least squares model.
    \item \textbf{SVR.} We used a Support Vector Regression model with a kernel containing a non-linear radial basis function. We also tuned the regularization parameter (traditionally denoted $C$) for these models. A larger value of $C$ optimizes for a decision boundary which separates the data more accurately, but with smaller margins, whereas a smaller value of $C$ optimizes for a larger margin around the decision boundary, but allows more misclassified points. When predicting rating, we found that a regularization parameter of 0.1 returned the best results. For conversation length,  a regularization parameter of 10 performed best.
    \item \textbf{Decision Tree and Random Forest.} When using the Decision Tree and Random Forest regression models, we ran experiments with maximum allowed depths of 5 and 10, as well as an unbounded maximum depth. For length, the unbounded depth trees consistently performed the best, but limiting the maximum depth to 5 performed best when predicting rating.
    \item \textbf{MLP.} When predicting conversation rating, we found that the performance of the Multi-layer Perceptron algorithm was poor when the hidden layers were large. So we used a model with just 5 hidden nodes in a single hidden layer. For  conversation length, we raised the maximum number of training epochs to 1000 and used a structure consisting of two hidden layers, the first of which contained 100 hidden nodes, the second containing 50 hidden nodes. We also tried single hidden layer models with 50, 100, and 150 hidden nodes, but these alternatives performed worse. In all cases, the ``reLu" activation function was used at each hidden node.
    \item \textbf{DistilBERT.} DistilBERT is a lightweight Transformer\cite{vaswani2017attention} model based on the BERT-base\cite{devlin2018bert} model. DistilBERT is trained by distilling BERT base so that it is smaller, faster, and cheaper, yet has only marginally lower performance. Compared to BERT, DistilBERT has 40\% less parameters, runs 60\% faster, and achieves 95\% of BERT's performance on the GLUE\cite{wang2018glue} language understanding benchmark. We initialize DistilBERT using the pre-trained weights available through Huggingface Transformers.\footnote{\url{github.com/huggingface/transformers}} 
\end{itemize}

\section{Results}
\label{results-sec}

We train each regression model using the features described in Section~\ref{methods-sec}, and report mean squared error (MSE), the coefficient of determination ($R^2$), and the Pearson correlation ($r$) for each model below.

\begin{wraptable}{r}{3.05in}
    \centering
    \small
    \vspace{-.25in}
    \begin{tabular}{|m{1.4in}|m{0.45in}|m{0.45in}|m{0.45in}|}
    \hline
        Model & MSE & $R^{2}$ & $r$\\\hline \hline
        SVR & 1.799 & 0.025 & 0.252**\\ \hline
        Decision Tree & 1.767 & 0.042 & 0.210** \\\hline
        Random Forest & 1.743 & 0.056 & 0.237**\\\hline
        Multi-layer Perceptron & 1.726 & 0.065 & 0.258** \\\hline
        Least Squares Linear & 1.709 & 0.074 & 0.272** \\\hline
        DistilBERT Transformer & 1.597 & {\bf 0.135} & 0.370** \\\hline
    \end{tabular}
    \caption{Results of regression models trained to predict the user provided ratings. ** indicates significant $r$ correlation at $p\leq .01$.}
    \label{tab:rating-results}
     \vspace{-.25in}
\end{wraptable}\noindent{\bf Predicting Ratings.} Table \ref{tab:rating-results} shows for each model, the mean squared error (MSE) between the model's predicted ratings and the ground-truth ratings, and the corresponding $R^2$ and $r$ values. The DistilBERT model yields the lowest MSE and the highest $R^2$ with an MSE of 1.597 and an $R^2$ of 0.135. While all the models'  $r$ values are significant, these model fits are  too low to be used to produce additional silver training data from unrated conversations. Given these values, we decided not to pursue training with Athena independent features. We plan to explore other approaches in future work.

\begin{wrapfigure}{r}{1.25in}
\vspace{-.2in}
\includegraphics[width=1.25in]{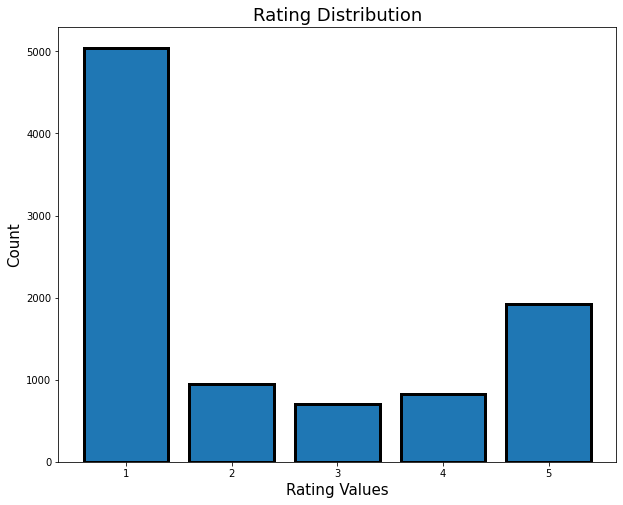} 
\includegraphics[width=1.25in]{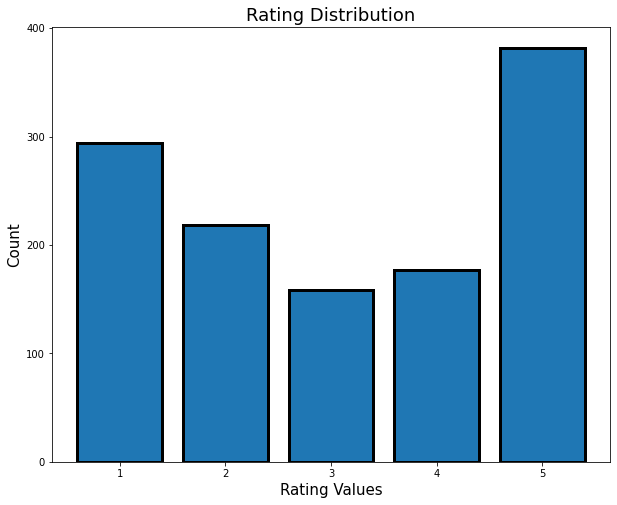} 
\caption{Distribution of Ratings for conversations of lengths$<$5, and those of  lengths 3\&4.} \label{short-ratings-fig}
\vspace{-.2in}
\end{wrapfigure}

We found it surprising that user ratings, one of the main evaluation criteria for the Alexa Prize, should be so challenging to predict, even given thousands of training examples \cite{shalyminov2018neural}. One reason might be subjectivity in  user ratings \cite{Venkatesh2017}. In previous work,  a controlled study with UCSC undergrads showed that  user personality and  user expectations of a spoken SocialBot, affected ratings \cite{bowden2019cui}. Athena ratings for conversations of lengths$<$5 also  provides evidence for user subjectivity, given that we believe that these conversations are too short to justify any rating at all.  
The top plot in Figure~\ref{short-ratings-fig} shows that most  users rate conversations of length $<$ 5  with a 1 but there is large variance ($\sigma^2 = $ 2.51), while the variance  for the whole training set is 1.85. The 5000 conversations of length 1, which only consisted of a greeting, were all rated 1.
Figure~\ref{short-convs-fig} showed short conversations of length 3 and 5. The variance in ratings for conversations of lengths 3\&4, at the bottom of Figure~\ref{short-ratings-fig}, clearly suggests an effect of user personality or subjectivity.   This is why we filter out these conversations from our training data. In other settings, it would be possible to normalize user ratings according to the ratings distribution for a particular user (rater) \cite{Callison09}, but only about 20\% of AP  users actually carry out multiple dialogues with Athena. While Alexa users may talk to an AP  system repeatedly, the random assignment of users to systems means an individual system is not likely to see the same user twice.

As a final point,  it is well
known that the performance of ASR  varies on an individual basis, and that  ASR error and ASR confidence scores are predictive of user satisfaction and conversation length. However AP  systems do not have access to acoustic properties or gold-standard transcriptions of user utterances, and therefore cannot model ASR error \cite{LitmanNAACL00,stoyanchev2012localized,walker2002automatically,Walker2000using,WKL00-jair}.

\noindent{\bf Predicting Conversation Length.}

We expect  better results for predicting conversation length, given the  correlation in Table~\ref{table:correlation-stats} of length with independent measures of user satisfaction (compliments). Recent work by Shalyminov et al \cite{shalyminov2018neural} also shows that   conversation length performs as well as user ratings  for reinforcement learning  in  AP conversations.  However, 
it is important when predicting conversation length, to ensure that features are represented in such as way to avoid indirectly encoding
 length. We thus only use features calculated as frequencies and medians. We also conduct ablation studies to show that individual features are not
indirectly encoding length.

\begin{figure}[h]
\vspace{-.1in}
\centering
\includegraphics[width=4.2in]{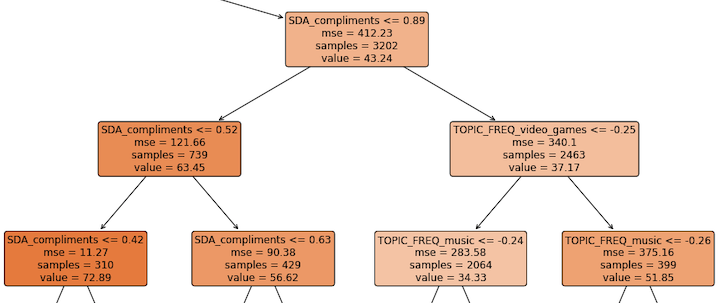}
\caption{Second, Third and Fourth layers of the Decision Tree predicting Conversation Length using Athena Dependent Features, with an R$^2$ model fit of 0.819 (see RHS of Row 2 of Table 4).}
\label{athena-dependent-tree-fig}
\end{figure}

Table \ref{tab:length-results} shows, for  each model and feature set,  surprisingly good  mean squared error (MSE),  and  $R^2$ and $r$ values.  A model based on the DistilBERT architecture achieved the lowest MSE at 8.20 for Athena  dependent features and the highest $R^2$ at 0.975. 
However, these models also are likely to rely on  many Athena independent features as suggested by 
the decision  tree model shown 
in Figure~\ref{athena-dependent-tree-fig}. The top node splits on the SDA\_compliments feature. The sub-branch of the tree we include in the figure shows that the SDA\_compliments feature is highly predictive of dialogue length, as the model chooses to split on that feature multiple times. The Athena dependent features representing topics such as video games and music are not selected for model splits until the third layer of the tree. This may also reflect
user subjectivity in their interests in different topics. 

\begin{table}[htb]
    \centering
    \begin{tabular}{|m{1.2in}|m{0.5in}|m{0.5in}|m{0.5in}||m{0.5in}|m{0.5in}|m{0.5in}|}
    \hline
       & \multicolumn{3}{c||}{Athena Specific Features} & \multicolumn{3}{c|}{Athena Independent  Features} \\ 
        Model & MSE & $R^{2}$ & $r$ & MSE & $R^{2}$ & $r$\\\hline \hline
        Least Squares Linear & 95.82 & 0.684 & 0.827**  & 249.23 & 0.179 & 0.425**\\ \hline
        Decision Tree & 55.03 & 0.819 & 0.909** & 80.69 & 0.734 & 0.866**\\\hline
        Random Forest & 24.91 & 0.918 & 0.959** & 41.78 & {\bf 0.862} & 0.929** \\\hline
        SVR & 24.90 & 0.918 & 0.959** & 106.63 & 0.649 & 0.824** \\ \hline
        Multi-layer Perceptron & 18.38 & 0.939 & 0.970** & 54.24 & 0.821 & 0.907**\\\hline
        DistilBERT Transformer & 08.20 & {\bf 0.975} & 0.995** & 43.00 & 0.842 & 0.920** \\\hline
    \end{tabular}
    \caption{Results of  regression models trained to predict the length of the conversations for  Athena Dependent and Athena Independent features. A ** indicates a significant $r$ correlation at $p\leq .01$.}
    \label{tab:length-results}
\end{table}

The z-scores in Figure~\ref{z-scores-fig} suggest that length prediction should benefit from topic features. Some topics such as music may be popular in any system, but   topic performance is clearly Athena Dependent, reflecting whether large scale training data on that topic is present in  corpora such as Topical Chat \cite{gopalakrishnan2019topical,hedayatnia2020policy}, whether high quality topical data is in WikiData or sources like IGDB, and the human effort put into that topic. Thus, it is surprising that the RHS of Table~\ref{tab:length-results} shows that the Athena Independent features  achieve excellent performance, with an $R^2$ value of 0.862 using a Random Forest model.  Since  SDA features can be calculated
for any dialogue system, these results imply that the Athena Independent models are very general. 

\begin{figure}[h]
\vspace{-.25in}
\includegraphics[width=4.6in]{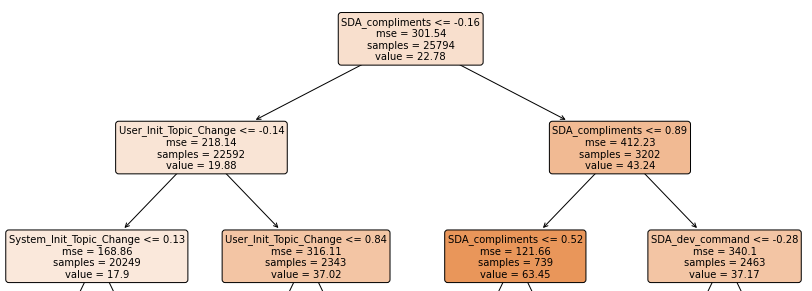}
\caption{Top 3 layers of a Decision Tree predicting Conversation Length using Athena Independent Features, with an R$^2$ model fit of 0.734 (LHS of Row 2 of Table 4).} 
\label{system-independent-tree-fig}
\vspace{-.25in}
\end{figure}
\begin{wraptable}{r}{3.25in}
    \centering
    \small
    \begin{tabular}{|m{1.9in}|m{0.34in}|m{0.34in}|m{0.4in}|}
    \hline
        Features Ablated & MSE & $R^{2}$ & $r$\\\hline \hline
        TOPIC\_DIST\_MEDIAN & 78.10 & 0.743 & 0.871**\\ \hline
        TOPIC\_FREQ\_music & 67.09 & 0.779 & 0.889** \\\hline
        TOPIC\_FREQ\_video\_games & 60.57 & 0.800 & 0.899**\\\hline
        SDA\_compliments & 50.38 & {\bf 0.834} & 0.917** \\\hline
        TOPIC\_FREQ\_movies & 61.54 & 0.797 & 0.898** \\\hline
        User\_Init, Sys\_Init & 72.26 & 0.762 &  0.880** \\\hline
        User\_Init, Sys\_Init, SDA\_compliments & 72.43&  0.761 & 0.880** \\\hline
    \end{tabular}
    \caption{Results of an ablation study  using the Decision Tree model. ** indicates significant $r$ correlation at $p\leq .01$.}
    \label{tab:ablation-results}
    \vspace{-.2in}
\end{wraptable}
Figure~\ref{system-independent-tree-fig} shows the top three layers of the Athena Independent decision tree model. After doing an initial split on the SDA\_compliments feature,
the model continues to rely on more general features. The LHS of the tree splits on the frequency of initiative dialogue acts that either select a topic or offer a menu of topics, in response to the user's request to change the topic.  The RHS of the tree continues to split on the SDA\_compliments feature, but also uses the SDA\_dev\_command dialogue act.

\begin{figure}[h]
\vspace{-.1in}
\includegraphics[width=4.5in]{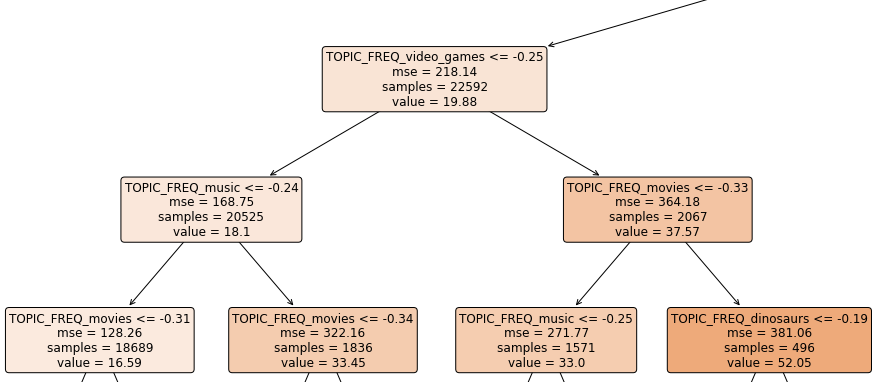}
\caption{Ablating Initiative and Compliments. 3 LHS layers of a Decision Tree predicting  Length using Athena Dependent  Features, with an R$^2$ model fit of .762 (See  Row 7 of Table 5). }
\label{ablation-tree-fig}
\vspace{-.2in}
\end{figure}

Table~\ref{tab:ablation-results} reports results for  an ablation study of  the top  Athena Dependent features according to their correlation with length.  We use the Decision Tree models and then examine the effect of ablating particular features. 
Figure~\ref{ablation-tree-fig} shows a decision tree after ablating User and System Initiative and  the SDA\_compliments feature.  This model achieves an R$^2$ of .762 using Athena Dependent topic features, indicating the topics that  lead to longer conversations. 



\noindent{\bf Predicting Length from Initial Sequences.} We  also explore  whether initial segments of the dialogue can  predict conversation length, as in  early work on {\sc problematic dialogue} prediction for task-oriented dialogue. Examination of our corpus suggests that some
users are adversarial, some conversations just go poorly from the start,
or some users suffer from particularly poor ASR due to dialect or native language  \cite{stoyanchev2012localized}. 
Such predictions can be added  to
the dialogue state table and used for reinforcement learning\cite{HastieetalACL02,LWRGL99,WLHWG02}. 
We experiment with initial segments
of length 10 and 15, given that the median conversations length  is 17 exchanges. 

 \begin{table}[htb]
    \centering
    \begin{tabular}{|m{1.1in}|m{0.5in}|m{0.5in}|m{0.5in}||m{0.5in}|m{0.5in}|m{0.5in}|}
    \hline
     & \multicolumn{3}{c||}{First 10 exchanges} & \multicolumn{3}{c|}{First 15 exchanges} \\\hline
      Model    & MSE & $R^{2}$ & $r$ & MSE & $R^{2}$ & $r$\\\hline \hline
          
        Least Squares Linear & 0.211 & 0.154 & 0.394** & 0.149 & 0.402 & 0.634**\\\hline
        Decision Tree & 0.226 & 0.095 & 0.508** & 0.086 & 0.656 & 0.828**\\\hline
        SVR & 0.192 & 0.228 & 0.549** & 0.100 & 0.597 & 0.775**\\\hline
        Random Forest & 0.159 & {\bf 0.361} & 0.612** & 0.047 & {\bf 0.812} & 0.902**\\\hline
        Multi-layer Perceptron & 0.174 & 0.301 & 0.579** & 0.081 & 0.676 & 0.827**\\\hline
    \end{tabular}
    \caption{Results of logistic regression models trained to predict  whether or not  the conversation length would be less or greater than the median. A  ** indicates a significant correlation at $p \leq .01$. }
    \label{tab:above-or-below-median-length-results-10-and-15}
    
\end{table}   

We first trained and tested logistic regression  by binning conversations lengths into two bins, one for  lengths less than the median, and another for  lengths greater than or equal to the median, with results in Table~\ref{tab:above-or-below-median-length-results-10-and-15}.
The $R^2$ of the Random Forest model after seeing 15 turns of conversation is 0.812, only 0.10 less than a  model that has seen the whole conversation, resulting in an excellent model for predicting whether the
conversation is likely to end in the next few turns.

\begin{table}[htb]
    \centering
     \begin{tabular}{|m{1.1in}|m{0.5in}|m{0.5in}|m{0.5in}||m{0.5in}|m{0.5in}|m{0.5in}|}
    \hline
   & \multicolumn{3}{c||}{First 10 exchanges} & \multicolumn{3}{c|}{First 15 exchanges} \\\hline 
      Model    & MSE & $R^{2}$ & $r$ & MSE & $R^{2}$ & $r$ \\ \hline \hline
        Least Squares Linear & 2.586 & 0.144 & 0.381** & 2.186 & 0.276 & 0.526**\\\hline
        Decision Tree & 3.159 & -0.045 & 0.439** & 2.768 & 0.084 & 0.554**\\\hline
        SVR  & 2.430 & 0.196 & 0.510** & 2.035 & 0.326 & 0.620**\\\hline
        Random Forest & 2.168 & {\bf 0.283}  & 0.556** & 1.645 & {\bf 0.456} & 0.683**\\\hline
        Multi-layer Perceptron & 2.269 & 0.249 & 0.536** & 2.035 & 0.326 & 0.612** \\\hline
    \end{tabular}
    \caption{Results of regression models trained to predict conversation length bins of size 10. A ** indicates a significant correlation at $p \leq .01$.}
    \label{tab:binned-length-results-10-15}
\end{table}

We then experiment with conversation length binned into  increments of 10 exchanges up to 70, with one final bin for conversations that were 70 exchanges or longer. These models use the same features as those in Table~\ref{tab:above-or-below-median-length-results-10-and-15}, with results in Table~\ref{tab:binned-length-results-10-15}. 
The best performing Random Forest model after 15 turns of conversation results in an R$2$ of 0.456. Thus it performs fairly well at  predicting how long the conversation will be in terms of 10 exchange chunks. Additional features and further tuning could easily lead to even better results.

\section{Conclusion and Future Work}
\label{conc-sec}

Our initial aim was to create a model that could predict the evaluation  metrics
that  AP  systems optimize for, namely dialogues that are  ``long and engaging'',
in the style of work on PARADISE and Interaction Quality \cite{Walkeretal97,ultes2016analysis}.
The setting of the AP  supports the collection of thousands of user ratings at the end of a conversation: these should be a valuable indication of how engaging a dialogue is. We created predictive 
models of the ratings  using Athena dependent dialogue features, but the best $R^2$ value was 0.139. We then showed that the variance in ratings of dialogues that are too short to merit a rating, shows  that there
is great deal of subjectivity in these ratings \cite{Venkatesh2017}.
In other settings, this subjectivity could be accounted for by normalizing ratings  for each user \cite{Callison09}, but only about 20\% of AP  users engage in multiple dialogues with Athena.  Other work on AP systems modelled  user individual differences such as Big Five personality, and user propensity to take the initiative \cite{fang2018sounding18,chen2018gunrock}. Athena models user topical preferences already, and e.g. makes inferences
that interest in a sport like hiking may indicate an interest in a topic such as nature, or that interest in a topic like food may indicate interest in a topic such as  nutrition. It seems clear that further work on personalized models of the user could be fruitful for better predicting user ratings \cite{stoyanchev2009lexical}.

However, 
the length metric is less subjective because users are volunteers, and only converse at length if they are engaged in the conversation.  Results
for predicting conversation length with a DistilBert model achieves an R$^2$ of  0.975,
using Athena Dependent features. Results for predicting length  with Athena Independent features are also excellent with an R$^2$ of 0.862 
for a Random Forest model. 

We also train models to predict length based on initial sequences of the dialogue of lengths 10 and 15. This ability is  important for using
length prediction in real-time to affect a dialogue system's behavior. 
The best  R$^2$ is 0.34
for a model that only had access to the first 10 exchanges of a conversation,
and the best  R$^2$ is 0.740
for a model that had access to the first 15 exchanges.

We believe that better results are possible, with additional features representing
conversational behaviors. 
Features such as user response time might indicate confusion or a lack of interest, while our previous work showed that system response time directly affects user ratings \cite{harrison2020athena}. Another speech feature missing from our models are ASR confidence scores and actual ASR error rates \cite{stoyanchev2012localized}. 
In future work, we will use the length predictor in the representation of dialogue state and condition the dialogue policy on predicted length.  We would also like  to  test our  model on other AP  systems and conversational agents.

\section{Acknowledgements}
The work in this paper was supported by an NSF Grant \#2019805 to  AI Institute for Student AI Teaming (UCSC subcontract), NSF IIS   \#1748056, and an  Amazon  Alexa Prize grant. Thank you to the rest of the team for their  work on Athena, and to folks at Amazon AI for feedback on an earlier version of this paper. 


\end{document}